\documentclass[preprint,12pt,authoryear]{elsarticle}

\usepackage{amssymb}
\usepackage{amsmath}
\usepackage{hyperref}
\usepackage{caption}

\usepackage{lineno}

\journal{Elsevier}

\begin{document}

\begin{frontmatter}



\title{Enhancing Pedestrian Trajectory Prediction with Crowd Trip Information} 



\author[a]{Rei Tamaru} 
\ead{tamaru@wisc.edu}
\author[a]{Pei Li\corref{cor1}}
\ead{pei.li@wisc.edu}
\cortext[cor1]{Corresponding author}
\author[a]{Bin Ran}
\ead{bran@engr.wisc.edu}
\affiliation[a]{organization={Department of Civil and Environmental Engineering, University of Wisconsin-Madison}, 
addressline = {1415 Engineering
Dr, Madison, 53706, WI, United States}}
\begin{abstract}
Pedestrian trajectory prediction is essential for various applications in active traffic management, urban planning, traffic control, crowd management, and autonomous driving, aiming to enhance traffic safety and efficiency. Accurately predicting pedestrian trajectories requires a deep understanding of individual behaviors, social interactions, and road environments. Existing studies have developed various models to capture the influence of social interactions and road conditions on pedestrian trajectories. However, these approaches are limited by the lack of a comprehensive view of social interactions and road environments. To address these limitations and enhance the accuracy of pedestrian trajectory prediction, we propose a novel approach incorporating trip information as a new modality into pedestrian trajectory models. We propose RNTransformer, a generic model that utilizes crowd trip information to capture global information on social interactions. We incorporated RNTransformer with various socially aware local pedestrian trajectory prediction models to demonstrate its performance. Specifically, by leveraging a pre-trained RNTransformer when training different pedestrian trajectory prediction models, we observed improvements in performance metrics: a 1.3/2.2\% enhancement in ADE/FDE on Social-LSTM, a 6.5/28.4\% improvement on Social-STGCNN, and an 8.6/4.3\% improvement on S-Implicit. Evaluation results demonstrate that RNTransformer significantly enhances the accuracy of various pedestrian trajectory prediction models across multiple datasets. Further investigation reveals that the RNTransformer effectively guides local models to more accurate directions due to the consideration of global information. By exploring crowd behavior within the road network, our approach shows great promise in improving pedestrian safety through accurate trajectory predictions.
\end{abstract}








\end{frontmatter}

\begin{figure*}[h]
  \centering
  \includegraphics[width=\textwidth]{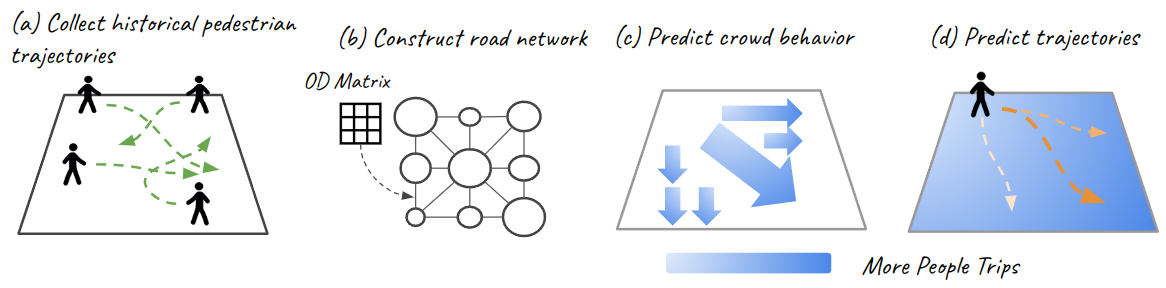}
  \caption{The concept of RNTransformer: pedestrian intention prediction with trip modality.}\label{fig:concept}
\end{figure*}

\section{Introduction}

Pedestrian trajectory prediction is essential in various applications that aim to improve traffic safety and efficiency. For instance, predicting pedestrian trajectories can be embedded in an assistive system in a vehicle or traffic control devices that alert drivers if pedestrians are likely to cross their paths, thus preventing potential conflicts and accidents \citep{mogelmose2015trajectory}. Similarly, autonomous vehicles can leverage these predictions to make safer decision-making, enhancing overall road safety \citep{li2020socially,zhang2020pedestrian,zhang2022adversarial}. Moreover, accurate future pedestrian trajectories are important for crowd management systems, helping to ensure safe and secure public spaces by anticipating and mitigating potential issues related to road users.

Pedestrian trajectories are affected by various factors, including individual characteristics, road environments, and social interactions. In particular, social interactions are crucial as pedestrians constantly adjust their trajectories based on their interactions with other road users and objects. The studies of socially aware models have investigated to understand the social interactions between pedestrians and other road users or objects. For example, Hilbing's social force model defines social interactions using social forces. Pedestrian motion is described as if individuals are subjected to social forces, including acceleration towards the desired velocity, a reflective (territorial) effect that maintains a social distance with other pedestrians, and an attractive effect from other road users or objects \citep{Helbing1993,hall1995handbook}. Early socially aware models represented these social interactions explicitly in the matrices or convolutional modules \citep{alahi2016social, Xue2018, pfeiffer2018data, nikhil2018}. More recent advances in deep learning have led to the use of recurrent modules \citep{Zhang2019, song2020pedestrian}, graph neural networks to reason the interaction \citep{Huang2019, Mohamed2020, mohamed2022social}, and attention modules \citep{makansi2020multimodal, yu2020spatio} to understand the social attributes. Despite these advancements, many existing methods primarily consider relative distances between pedestrians and other road users and often overlook a more comprehensive global perspective.

Moreover, pedestrian intentions are studied to understand long-term behavior and predict pedestrian trajectory, which in turn helps reveal the road environments's impact on their behavior~ \citep{Ikeda2013, Tran_2021_WACV, girase2021loki}. Specifically, multimodal learning has merged as a key approach for predicting pedestrian intentions, capturing the scene context, and enhancing trajectory prediction accuracy \citep{mangalam2021, lv2023}. Recent research has explored visual modalities, such as semantic priors \citep{rasouli2019pie, malla2020titan} and scene segmentation \citep{makansi2020multimodal, yang2022}, to guide future trajectory predictions. Additionally, the optical flow has been used to extract motion features \citep{styles2020, yin2021}. However, the effectiveness of visual modality can be limited by factors such as restricted perception perspectives and irrelevant motion in image sequences. Given the necessity for a deep understanding of contextual information in predicting pedestrian behavior, we propose \textit{trip modality} as a new modality that incorporates crowd trip information. 


In summary, further enhancing pedestrian trajectory prediction requires a deeper understanding of interactions among pedestrians, road users, and other objects while incorporating global information. Moreover, robust modalities should be introduced to help understand the impact of the road environments on pedestrian trajectory. In this paper, we introduce the Road Network Transformer (RNTransformer), a model that integrates pedestrian intention with the representations of crowd behavior, providing a global context of trip information to the pedestrian trajectory prediction model. Figure \ref{fig:concept} illustrates our concept of RNTransformer. The process begins with collecting historical pedestrian trajectories from the dataset to construct the road network. This network then facilitates the sampling of pedestrian goals, which the RNTransformer uses to predict crowd behavior. Finally, these predicted features are cooperatively fused to predict individual pedestrian intentions. By utilizing a pre-trained RNTransformer during the training process of the pedestrian trajectory prediction model, we achieved a 1.3/2.2\% improvement on Social-LSTM, 6.5/28.4\% improvement on Social-STGCNN, and 8.6/4.3\% improvement on S-Implicit in ADE and FDE, respectively. Through extensive experimentation, our model demonstrates enhanced prediction accuracy across various baseline pedestrian trajectory prediction models and shows that RNTransformer significantly contributes to accurately sampling pedestrian goals. The contributions of this paper are summarized as follows:

\begin{enumerate}
    \item We propose trip information as a new modality, representing global features in a scene, enhancing the understanding of pedestrian behavior beyond traditional methods.

    \item We integrate the pre-trained RNTransformer into trajectory prediction models, improving the accuracy of pedestrian trajectory prediction.

    \item We design a method for constructing road networks that learn crowd behavior by segmenting the scene into multiple grids and aggregating pedestrian counts over time, capturing spatial and temporal dependencies.
    
    \item The RNTransformer shows significant improvements in prediction accuracy for various baseline models, validated through extensive experiments on a variety of datasets. The code is available at \url{https://github.com/raynbowy23/RNTransformer}.
    
\end{enumerate}

\section{Literature Review}

The approaches to understanding pedestrian trajectory are generally categorized into expert-based, data-driven methods, or hybrid approaches of them \citep{golchoubian2023pedestrian}. Expert-based models rely on hand-crafted rules to simulate pedestrian decision-making, delineating precise human kinetics \citep{anvari2015}. In contrast, data-driven approaches, particularly those using deep learning, better capture the complexity of real-world scenarios \citep{girase2021loki, rasouli2019pie}. The hybrid of these methodologies offers a balanced approach, leveraging environmental understanding while imposing constraints on pedestrian behavior through established models, such as the social force model \citep{Helbing1993}, thus improving the overall prediction accuracy and performance \citep{rudenko2020human}. This balance between structure and flexibility has proven to be effective in adjusting for the unpredictability of pedestrian behavior in dynamic environments.

\subsection{Socially Aware Pedestrian Trajectory Prediction}
Social interactions between pedestrians and other road users or objects are a key factor in predicting future trajectories. Such interactions are obtained by socially aware models. \citet{alahi2016social} firstly integrated social interactions, as derived from the social force model, into a deep learning framework. They introduced a social pooling layer to represent spatial relationships, allowing spatially adjacent long short-term memory networks (LSTMs) to share information. Building on this, \citet{bisagno2018group} proposed Group LSTM, which clustered individuals by their locations before predicting their trajectories. This work highlighted the importance of acknowledging group dynamics within pedestrian interactions. Expanding on this concept, \citet{gupta2018social} introduced social interactions into a generative adversarial network (GAN) framework. The proposed model features a generator, a pooling module, and a discriminator. The generator predicted the pedestrian trajectories, while the discriminator, by leveraging subtle social interaction rules, classified socially unacceptable trajectories (e.g., those causing collisions) as "fake." The model used LSTMs as its encoder and decoder, with a pooling layer aggregating information across pedestrians. This designed pooling module considered interactions between all pedestrians, while the pooling module in \cite{alahi2016social} only considered people inside the grid it has defined, thereby improving predictive accuracy over Social-LSTM. 

Later studies further advanced the prediction accuracy by integrating more complex interaction models. \citet{song2020pedestrian} developed a convolutional LSTM model that accounted for both spatial and temporal interactions using multi-channel tensors, achieving significant improvements in predictive accuracy. Similarly, \citet{Mohamed2020} proposed Social-STGCNN that modeled social interactions among pedestrians as a graph. A $20\%$ improvement in terms of prediction accuracy has been achieved over the state-of-the-art models. These models illustrate how embedding social interactions accelerates both training and performance, demonstrating the effectiveness of modeling pedestrian dynamics as a core component of trajectory prediction.

\subsection{Pedestrian Intention Prediction}
Extended studies on crowd dynamics, such as those conducted by \citet{helbing2000} on panic situations, have shown that crowd psychology influences individual movements. Pedestrian intention prediction aims to estimate future pedestrian behavior from a macroscopic viewpoint, accounting for multimodality to handle inherent uncertainty in pedestrian movements. Several goal-based models have addressed this challenge, including preset goals \citep{Tran_2021_WACV} and multimodal goal probability distributions \citep{Ikeda2013, mangalam2021, girase2021loki}. \citet{Tran_2021_WACV} proposed a matching that aligns provided goals with observed trajectories, leveraging recurrent modules. The output representations are concurrently used to guide pedestrians toward their intended destination. Similarly, the sub-goal model proposed by \citet{Ikeda2013} posits that pedestrians head towards a series of sub-goals en route to the final destination, with each sub-goal inferred from past trajectories and human kinetics. \citet{girase2021loki} developed a goal estimation model with a conditional variational autoencoder, demonstrating that reasoning about goals in a recurrent manner can significantly improve trajectory predictions. 

Although these goal-based approaches provide valuable insights, our proposed RNTransformer extends this line of inquiry by capturing global scene representations constructed with historical trajectories. Similar to the grid-based approach described by \citet{gu2021} and \citet{Derrow_Pinion_2021}, RNTransformer anticipates that pedestrian goals can be depicted in a grid-based manner, which accommodates a wide range of behaviors such as slowing down or stopping.

\subsection{Multimodal Learning on Trajectory Prediction}
The prediction of pedestrian trajectories requires multimodal information to accurately capture the dynamic motion of pedestrians, as well as their interactions with other road users. Visual modalities have been extensively studied in existing research. For example, \citet{rasouli2019pie} estimated the crossing intention of pedestrians using an LSTM-based model from image data. Similarly, \citet{malla2020titan} employed a deep learning model to extract pedestrian actions, such as standing, jumping, running, and walking, from image sequences. These actions were used as prior information to provide meaningful interactions and important cues for predicting pedestrian trajectories. These studies leveraging RGB image data offer substantial insights, enabling reasoning about not only pedestrian movement but also their surroundings. 

Further advancements have focused on refining visual cues. For instance, \citet{yin2021} integrated the optical flow into a transformer-based model, estimating dynamic motion information between pedestrians and ego vehicles, while \citet{makansi2020multimodal} applied a semantic segmentation technique to infer static scene semantics (e.g., road, sidewalk) from image sequences, using this semantic information to predict future pedestrian locations. Y-Net from \citet{mangalam2021} predicts future trajectories by estimating goals with segmentation maps and waypoint distributions and heatmap inputs. However, the effectiveness of visual modality can be limited by factors such as restricted perception perspectives and irrelevant motion in the image sequences.

Our proposed trip modality, in contrast, addresses these challenges as it considers both the dynamic motion of crowds and the constraints of road facilities. Unlike visual modalities, which focus on isolated movements, the trip modality captures a more holistic view of crowd behavior within the road network. Moreover, the modularity of the RNTransformer framework allows seamless integration with any model along with other modalities, enabling more accurate and contextualized trajectory predictions.

\section{Methods}
Our proposed model considers trajectory predictions with hierarchical structures. The model has two components, including local and global prediction models. The study area is first divided into various nodes. The local model predicts individual pedestrian trajectories within each node. Differently, the global model predicts crowd behaviors with a sparse spatio-temporal module. In this study, we introduce a road network model that creates potential walkable areas with empirical pedestrian trajectories and geometrical information. Predictions from the RNTransformer are fed into the local model to predict the trajectories of pedestrians.

\begin{figure*}[t]
  \centering
  \includegraphics[width=0.9\textwidth]{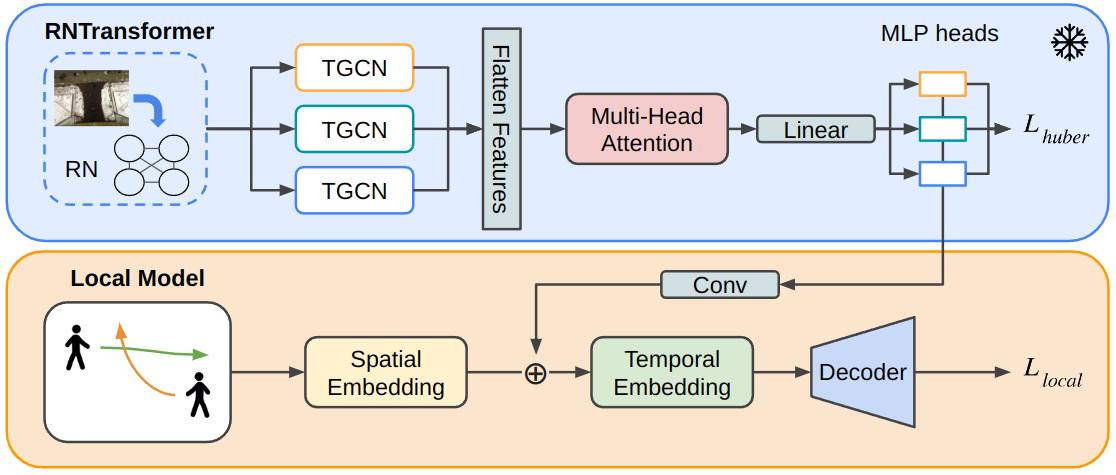}
  \caption{The architecture of the proposed model.}\label{fig:model_architecture}
\end{figure*}

Figure \ref{fig:model_architecture} illustrates the overall architecture of our proposed model. In our methodology, we first construct a representation of the target scene in the road network along the RNTransformer, which is utilized to train and evaluate crowd behavior in node locations within the network. Subsequently, the estimated latent vector is frozen and propagated through the convolution layer and embedded in the local prediction model. By employing multiple temporal layers, the model generates predictions for the variables necessary to describe the bivariate Gaussian distribution, from which a single sample of future pedestrian trajectories is derived.

\subsection{Model Architecture}
\subsubsection{Trajectory Prediction Model}
Socially aware pedestrian trajectory prediction models consider the relationships between individual pedestrians and their surroundings. These models construct adjacency matrices around each pedestrian, which are calculated as social tensors \citep{alahi2016social} or graph neural networks (GNNs) with nodes representing pedestrian locations and edges representing their interactions \citep{Mohamed2020, mohamed2022social}. These socially aware models typically combine the spatial awareness modules, such as convolutional neural networks (CNNs) or graph convolutional networks (GCNs), with the temporal awareness modules like time-extrapolator convolution neural networks (TXP-CNNs), recurrent neural networks (RNNs), and LSTMs. By stacking several layers of these spatial and temporal modules, the models can capture complex interactions and dependencies in pedestrian trajectories.

In the general setup for socially aware pedestrian trajectory prediction, the location of the $i^{th}$ pedestrian at time $t$ is denoted by $p^t_i = (x^t_i, y^t_i)$, where $t \in \{ 1, \ldots , T \subseteq \mathbb{N} \}$, and $x$ and $y$ represent the coordinates of pedestrian $i$ in $I$. The objective is to utilize historical trajectories from $0$ to $T$ to predict future trajectories from $T$ to $T + L$. Given the inherent uncertainty in future trajectories, they are traditionally sampled from a stochastic distribution. Specifically, $(x^t_i, y^t_i)$ are variables randomly sampled from a bivariate Gaussian distribution such that $p^t_i \sim \mathcal{N}(\mu^t_i, \sigma^t_i, \rho^t_i)$, where $\mu^t_i$ denotes the mean, $\sigma^t_i$ is the variance, and $\rho^t_i$ denotes the correlation of the distribution. Similarly, predicted trajectories are represented as $\hat p^t_i \sim \mathcal{N}(\hat \mu^t_i, \hat \sigma^t_i, \hat \rho^t_i)$ where the model calculates each of these variables.

Socially aware models consist of socially aware spatial modules and temporal reasoning modules. The spatial module transforms input features into hidden embedding using techniques such as a simple encoder \citep{alahi2016social} or graph convolution \citep{Mohamed2020, mohamed2022social} with the below process.

\begin{align}
v^i(l+1) = \sigma (\frac{1}{\Omega} \sum_{v^{j(l)} \in B(v^{i(l)})} \psi \cdot w(v^{i(l)}, v^{j(l)}))
\end{align} \\
where $\frac{1}{\Omega}$ is a normalization term. The set of neighboring vertices $B(v^i) = \{v^j | d(v^i, v^j) \leq D\}$ includes all vertices $v^j$ within a threshold distance $D$ from $v^i$, where $d(v^i, v^j)$ represents the shortest path length between $v^i$ and $v^j$. The function $\psi$ is a sampling function, and the kernel function $w(\cdot)$ captures the relationships (e.g., the inverse of the Euclidean distances) between nodes $v^i$ and $v^j$. Specifically, $w(v^{i(l)}, v^{j(l)}) = \frac{1}{||v^{i(l)}_t - v^{j(l)}_t||_2}, ||v^{i(l)} - v^{j(l)}_t||_2 \neq 0$; otherwise, $w$ is 0. 

These spatial representations are then input to the temporal prediction module, which can utilize CNN, GRU, LSTM, or other models to understand temporal reasoning. The residual connection, as shown in Figure \ref{fig:model_architecture}, allows past information to be retained in the predicted frame. In the local model, the task is to predict trajectories for $L$ future steps based on $n$ historical steps. Denote $V_t \in \mathbb{R}^{N \times I_t}$ as the node features at time $t$ and $G_p$ as the graph or adjacency matrix representing social interactions between pedestrians. The pedestrian trajectory prediction task can be represented as follows.

\begin{align}
    [V_{t+1}, \ldots , V_{t+L}] = f(G_p; V_{t_n}, ..., V_{t-1}, V_t)
\end{align}

\subsubsection{Road Network Construction}

\begin{figure*}[t]
  \centering
  \includegraphics[width=1\textwidth]{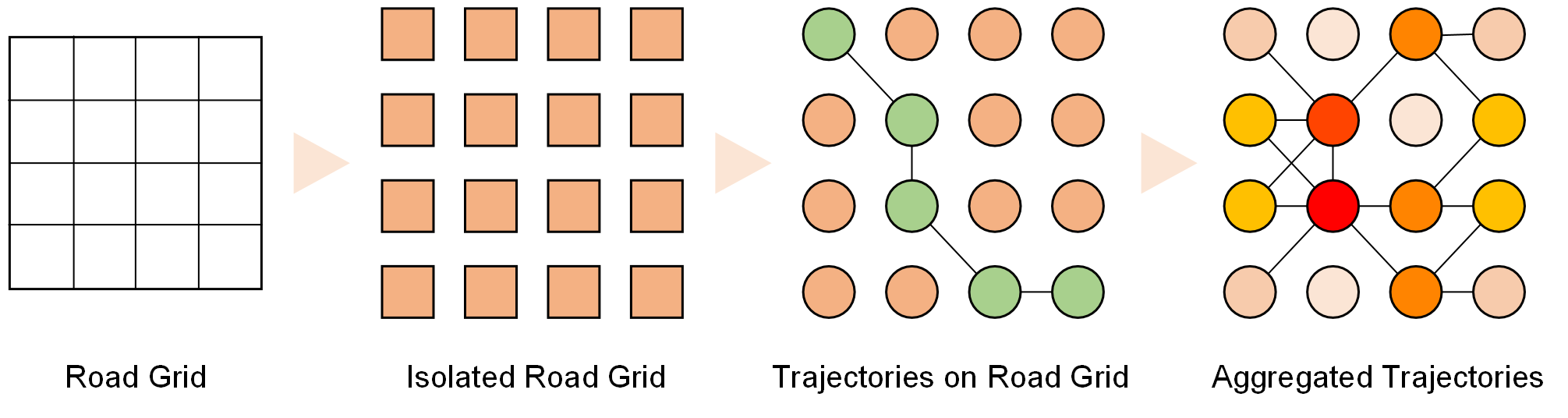}
  \caption{Road network construction process. The color of each node represents its degrees of aggregated trajectories. Green indicates the individual pedestrian trajectory and thicker-colored nodes have higher degrees.} \label{fig:road_network}
\end{figure*}

Inspired by the ETA prediction on road networks using supersegment construction \citep{Derrow_Pinion_2021}, we construct segmented grids on the target roadway to predict crowd behavior. We model the roadway as a geometrical graph with nodes divided by a preset number of grids for crowd movements and edges connecting each node. The road network $\mathbb{R}^{n \times m}$ is constructed based on historical pedestrian trajectories as illustrated in Figure \ref{fig:road_network}.

\begin{align} \label{eq:3.1}
    (n, m) = (\frac{\max{R_x} - \min{R_x}}{gr}, \frac{\max{R_y} - \min{R_y}}{gr})
\end{align} \\
where $gr$ denotes the number of grids. Then we isolate the road grid and create nodes $v_i \in V$ associated with each grid. Each node represents the number of pedestrians in each time step, and the edges indicate transitions from the previous number of pedestrians. We used origin-destination relationships in this paper. The function $\phi$ maps the road grid to the road network, initializing the value on the road grid 0 when the point is out of the scene coordinates and otherwise is set to 1.

\begin{align} \label{eq:3.2}
\phi_i =
\begin{cases}
    1, & \text{if } \min(R) \leq p^t_i \leq \max(R) \\
    0, & \text{otherwise}
\end{cases}
\end{align} \\

\begin{align}\label{eq:3.3}
    v_i = \bigcup_{t \in T}\bigcup_{i \in I} \phi_i p^t_i
\end{align} \\

The separated trajectories of each road user are merged into a road network. Each node center is determined by calculating the mean of the node size, allowing for an accurate representation of pedestrian distribution and movement within the network.

\subsubsection{Road Network Transformer}
Our Road Network Transformer (RNTransformer) leverages a GNN to predict the number of pedestrians on each node for several time horizons. Multiple models trained on different horizons are combined to predict pedestrian intentions. The graph constructed in the previous section is denoted as $G = (V, E)$.

Our approach integrates spatial and temporal geometrical relations through trip inputs and employs concatenation to merge the results from different time horizons. The model consists of a GCN as the spatial module and a GRU as the temporal module \citep{bai2021}. The spatial GCN processes node attributes, such as the number of road users and node centers, while another GCN compresses spatial relationships among nodes based on pedestrian frequency.

\begin{align}
\begin{split}
    H_{t'+1} &= g(f(H_{t'}, A)) \\
             & \text{where}\  f(x, A) = \sigma(W h_i + W h_i^{m}), \quad \quad h^{m} = \sigma(W \sum_{j\in d_i} \frac{e_{i,j}}{\sqrt{d_i d_j}} h_j)
\end{split}
\end{align} \\
In this context, $W$ denotes the weights, $d_i$ represents the degree of nodes, $\sigma$ denotes the activation function, and $d_i$ considers the graph neighborhood within $m$ hop.

Given input sequences with time steps $t' \in \{1, 2, \ldots, T' \}$, the temporal model predicts the future number of pedestrians $\pi$ for $L'$ subsequent steps. The input graph sizes are varied depending on the prediction horizon, which is a hyperparameter, and these sizes are forwarded to the spatial models to extract node features. Each feature encodes road structure information and node-specific attributes. The aggregation function is defined as $g(\cdot)$.

The node embeddings spatial model is employed to extract node attributes, while the temporal model captures temporal dependencies through road representations. These representations are further encoded into latent vectors using multiple linear layers. The vectors are flattened to the same length and concatenated to a single embedding. We then apply a Feed-Forward Transformer and a latent layer to these vectors, which are decoded into prediction values with corresponding heads. This final output serves as a trip representation.

\subsection{Model Training} \label{sec:training}
During the training process, the RNTransformer is used as a pre-trained model, with its parameters frozen. We adopt multiple loss functions for both the local and global models to ensure optimal performance. By iterating over the road network dataset, the RNTransformer learns node-wise information and compares predicted results with the actual pedestrian transitions using the Huber Loss $L_{global}$, which is advantageous for handling outliers.

We integrate features extracted by the RNTransformer into the local model with a learnable weight, where each feature is concatenated to the spatial layer within the local model for each road user. The architecture comprises a single convolution layer and multiple linear layers to adapt to the dynamic size of the tensors, the quantity of which varies in accordance with the number of prediction horizons considered. The ReLU activation function is applied during this process.

The local model assesses the differences between predicted and ground truth trajectories sampled from a bivariate Gaussian distribution using a negative log-likelihood loss. To enhance training stability, we also incorporate L1 and L2 regularization terms. Each loss function is iteratively applied to the overall training process, as shown below:

\begin{align}
\begin{split}
    L &= \lambda_{huber} L_{huber} + \lambda_{local} L_{local} + \lambda_{L1} L1 + \lambda_{L2} L2 \\
      &= \lambda_{huber} \frac{1}{bs}\sum^{K}_{t'=T'} \sum^{I}_{i=0}0.5(p^{t'}_i - \hat p^{t'})^2 
      \\
      & \quad - \lambda_{local} \sum^{T_p}_{t=T} \sum^{I}_{i=0}\log(P((p^t_i|\hat \mu^t_i, \hat \sigma^t_i, \hat \rho^t_i), (p^t_i | \mu^t_i, \sigma^t_i, \rho^t_i)) \\
      & \quad + \lambda_{L1} \sum |w_i| + \lambda_{L2} \sum |w_i^2|
\end{split}
\end{align} \\
where $bs$ denotes the model batch size and $w$ represents the parameters. $\lambda_{huber}, \lambda_{local}, \lambda_{L1}$, and $\lambda_{L2}$ are hyperparameters that balance the contributions of each loss component.

\section{Experiments}
\subsection{Experiments Design}
For our experiments, we utilized the ETH Walking Pedestrians Dataset \citep{lerner2007crowds} and the UCY Crowds-by-Example Dataset \citep{pellegrini2009}. These datasets contain pedestrian trajectories captured at multiple locations, and we divided each dataset into training and evaluation subsets to ensure robust model validation. The trajectories were recorded at a frequency of 2.5 FPS. For our model, we used 8 frames (3.2 seconds) as input observations and predicted the subsequent 12 frames (4.8 seconds).

The road network was constructed based on the entire training dataset spanning all time steps and divided into $6 \times 6$ grids for our experiment. Our prediction model was trained using the observed sequences of pedestrian trajectories and validated against the target sequences. Initially, the pre-trained road network was trained and tested independently. We then selected the best-performing model to fuse its hidden features into the local prediction model. The RNTransformer was trained for 50 epochs optimized using SGD with a learning rate of $10^{-2}$ and a weight decaying rate of $10^{-3}$.

We sampled multiple trajectory probabilities from Gaussian distribution, with each parameter of the Gaussians determined by the model. The trajectory prediction model was trained along three socially aware baseline models: Social-LSTM (S-LSTM) \citep{alahi2016social}, Social-STGCNN (S-STGCNN) \citep{Mohamed2020}, and Social-Implicit (S-Implicit) \citep{mohamed2022social}. The S-LSTM was trained using the AdaGrad optimizer, while S-STGCNN and S-Implicit were trained with the SGD optimizer. We applied the basic loss functions (Section \ref{sec:training}), as well as the specific loss functions defined in the respective papers for each baseline model.

We used two metrics to evaluate the developed model: the Average Displacement Error (ADE) \citep{alahi2016social} (Equation \ref{eq:ade}) and the Final Displacement Error (FDE) \citep{pellegrini2009} (Equation \ref{eq:fde}). ADE measures the average L2 distance between the predicted and ground truth trajectories across all predicted time steps. FDE, on the other hand, is defined as the L2 distance between the predicted position at the final time step and the actual ground truth position at that time step. In summary, ADE evaluates the model's performance on the entire prediction horizon, while FDE assesses its performance at the final time step.

\begin{align} \label{eq:ade}
    \text{ADE} = \frac{\sum_{i\in I}\sum_{t\in T_p} ||\hat p^t_i - p^t_i||_2}{I \times T_p}
\end{align}

\begin{align} \label{eq:fde}
    \text{FDE} = \frac{\sum_{i\in I}||\hat p^t_i - p^t_i||_2}{I},\ t = T_p
\end{align}

\subsection{Experimental Result}

\begin{table*}[t]
\caption{Performance comparison of baseline models and models integrated with RNTransformer (+RN). Each model was iteratively tested five times, with the results averaged to account for random sampling variability (ADE/FDE).}

\centering
\resizebox{\textwidth}{!}{\begin{tabular}{lcccccc} 
    & \multicolumn{1}{c}{S-LSTM} 
    & \multicolumn{1}{c}{\textbf{S-LSTM + RN}}
    & \multicolumn{1}{c}{S-STGCNN}
    & \multicolumn{1}{c}{\textbf{S-STGCNN + RN}}
    & \multicolumn{1}{c}{S-Implicit} 
    & \multicolumn{1}{c}{\textbf{S-Implicit + RN}}
    \\
\hline
\rule{0pt}{3ex} ETH & 1.04/1.24 & 1.03/1.23 & 0.72/1.40 & 0.64/1.05 & 0.67/1.43 & 0.64/1.36 \\
\rule{0pt}{3ex} Hotel & 0.77/0.88 & 0.74/0.85 & 0.43/0.73 & 0.40/0.61 & 0.32/0.60 & 0.26/0.52 \\
\rule{0pt}{3ex} Univ & 0.67/0.74 & 0.68/0.73 & 0.49/0.86 & 0.44/0.71 & 0.30/0.57 & 0.30/0.57 \\
\rule{0pt}{3ex} Zara1 & 0.68/0.85 & 0.67/0.83 & 0.34/0.54 & 0.36/0.57 & 0.24/0.46 & 0.23/0.46 \\
\rule{0pt}{3ex} Zara2 & 0.60/0.73 & 0.59/0.73 & 0.32/0.50 & 0.31/0.48 & 0.21/0.40 & 0.21/0.41 \\ 
\hline
\rule{0pt}{3ex} Avg. & 0.75/0.89 & \textbf{0.74/0.87} & 0.46/0.81 & \textbf{0.43/0.68} & 0.35/0.69 & \textbf{0.33/0.66} \\ \hline 
\end{tabular}} \label{tab:perform_comp}
\end{table*}

The model's performance on test data was evaluated using ADE and FDE. Table \ref{tab:perform_comp} presents the normalized results for these metrics across various datasets. The performance improvements are evident when comparing the baseline models with their extended versions that incorporate the RNTransformer. For the S-LSTM, integrating the RNTransformer results in a slight improvement with $1.3\%$ in ADE and $2.2\%$ improvement in FDE. The S-STGCNN shows significant performance gains with the RNTransformer integration. The ADE improves by $6.5\%$, and the FDE scores a remarkable improvement of $28.4\%$. The S-Implicit benefits greatly from the RNTransformer, achieving $8.6\%$ improvement in ADE and $4.3\%$ in FDE. Among the models tested, the S-Implicit with RNTransformer exhibits the best overall performance for both ADE and FDE, highlighting the synergy between the implicit modeling of social interactions and the road network representation provided by the RNTransformer.

\begin{figure*}[h]
\centering
  \includegraphics[width=1\textwidth]{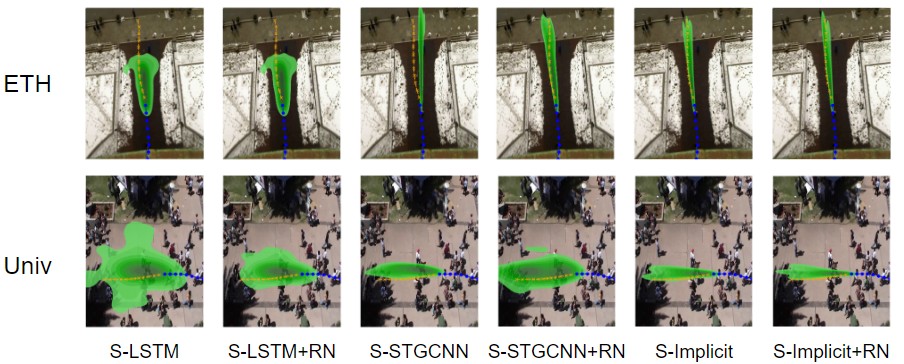}
  \caption{Visualization of the results applying baselines and our models to ETH and University (Univ) datasets. The blue line represents the observed trajectories, the orange line represents ground truth, and the green area indicates the predicted area.}\label{fig:qual}
\end{figure*}

Figure \ref{fig:qual} visualizes predicted pedestrian trajectories for two datasets: ETH and University (Univ) in the UCY dataset. The figure compares the performance of three baseline models, both with and without the integration of the RNTransformer. The possible predicted trajectories are represented in green areas, with warmer colors indicating areas of higher possibility. For the ETH dataset, the S-LSTM predicts pedestrian trajectories that generally follow the observed paths but show a wider spread. When the RNTransformer is integrated, there is a slight quantitative improvement, though the visual differences are minimal. The S-STGCNN also displays a relatively accurate prediction but with some dispersion in the predicted area. Adding the RNTransformer again refines these predictions, demonstrating a more concentrated trajectory that better matches the ground truth movement. The S-Implicit model exhibits a more accurate prediction compared to the other baseline models, with less dispersion in the predicted trajectories. Integrating the RNTransformer further enhances this accuracy, resulting in a very tight alignment with the ground truth trajectories.

For the Univ dataset, the S-LSTM model predicts trajectories that cover a broader area, indicating a degree of uncertainty in pedestrian movements. With the RNTransformer, the predicted trajectories become more concentrated, showing improved accuracy. The S-STGCNN model performs well, with predictions that closely follow the ground truth but still exhibit some spread. Integrating the RNTransformer results in a significant improvement, with predictions closely aligning with the actual pedestrian movements. The S-Implicit model provides the most accurate predictions among the baseline models, with a minimal spread in the trajectories. The addition of the RNTransformer further refines these predictions, demonstrating the highest accuracy and alignment with the ground truth.

Additionally, we can observe a multimodal distribution in the probability of the predicted trajectories with the S-Implicit. This suggests that the model captures multiple potential paths that pedestrians might take, reflecting the inherent uncertainty and variability in pedestrian behavior. However, when the RNTransformer is integrated, the mixture of these distributions becomes more apparent. The combined model effectively consolidates the multiple predicted paths, providing a more cohesive and accurate prediction that aligns well with the ground truth trajectories.

\begin{figure}[h]
\centering
  \includegraphics[width=\linewidth]{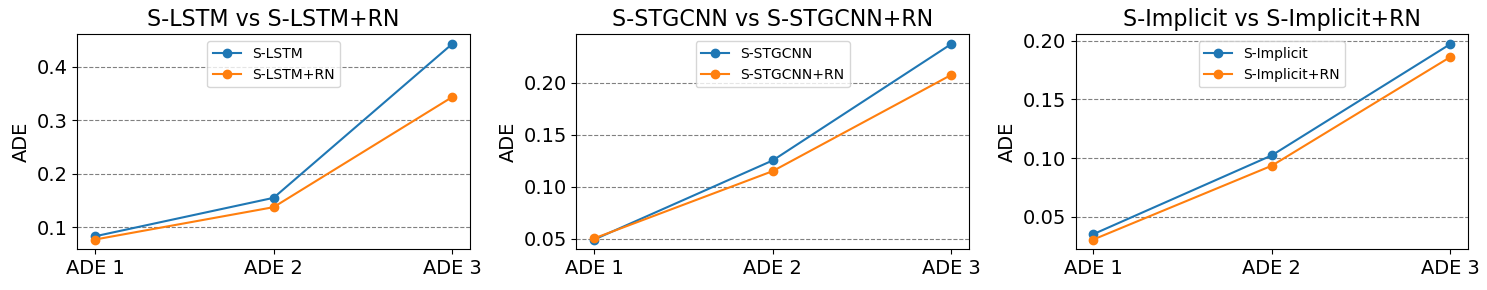}
  \caption{ADE at multiple segments. Each segmented ADE has four time steps.}\label{fig:ades}
\end{figure}

We also compared the segmented ADE. Since the FDE is an evaluation metric that only considers the coordinates at the final time step, we separated ADE into three segments and compared the dispersion of predicted trajectories in each segment. Figure \ref{fig:ades} shows the results on segmented ADE. For S-LSTM and S-STGCNN results, although their trends are similar in that ADE at early time steps, our model shows better ADE performance towards the later time steps. Specifically, the integration of the RNTransformer results in lower ADE values in the second and third segments, indicating more accurate long-term predictions. In the case of S-Implicit, the RNTransformer also improves performance with a similar tendency. This indicates that our model can provide a better goal to the local prediction model, enhancing accuracy over the entire prediction horizon.

\section{Discussion \& Future Works}
\begin{figure}[h]
    \centering
    \includegraphics[width=0.6\linewidth]{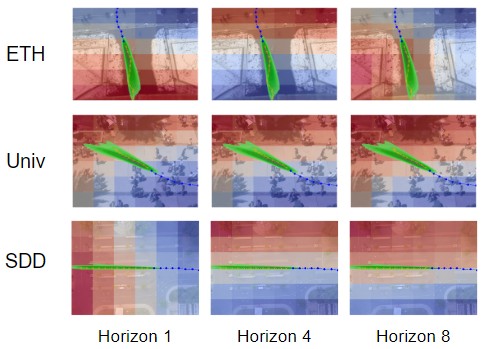}
    \caption{Visualization of overlay of crowd behavior prediction with different time horizons. A warmer color indicates more pedestrians compared to a colder color.}\label{fig:qual_rn}
\end{figure}


In the previous section, we observed that the trip modality enhances pedestrian trajectory prediction. We further investigated how RNTransformer integrates trip information into its network. Figure \ref{fig:qual_rn} visualizes the predicted pedestrian trajectories, and overlays heatmaps of RNTransformer prediction results across three different datasets over multiple time horizons. For the ETH dataset, RNTransformer's prediction of pedestrian density at the next time step (horizon 1) closely resembles that of the local trajectory model. However, the prediction at horizon 4 diverges from the actual pedestrian trajectory, and the horizon 8 prediction inaccurately forecasts pedestrian presence in the lower left area, which is out of bounds for the ETH dataset. In contrast, the University dataset results show that pedestrians tend to occupy the upper area consistently across all horizons, aligning with the local pedestrian trajectory model's results. Similar results are obtained using the SDD dataset as well. These qualitative findings indicate that RNTransformer contributes to pedestrian trajectory prediction by capturing crowd behavior, allowing the local trajectory model to select the most plausible results from each horizon.

Despite training RNTransformer on all five datasets, it occasionally predicted the number of pedestrians in unrealistic areas, such as in the ETH dataset. To further evaluate RNTransformer's contribution to the local model, we tested it on the Stanford Drone Dataset (SDD) \citep{robicquet2016learning}. We used a pre-trained RNTransformer, trained on the Nexus scenario, alongside whole models trained on multiple scenarios. Quantitatively, the S-Implicit achieved 0.48 in ADE and 0.90 in FDE. When integrated with the RNTransformer, the model improved to 0.44 in ADE and 0.81 in FDE, representing an 8\% and a 10\% improvement in ADE and FDE, respectively. As shown in Figure \ref{fig:qual_rn}, the SDD results at time horizon 1 predict the number of pedestrians in the left area, while predictions at later horizons forecast the number of pedestrians in the upper area. This demonstrates that RNTransformer offers plausible directional predictions for pedestrians.

\begin{figure}[h]
    \centering
    \includegraphics[width=0.9\linewidth]{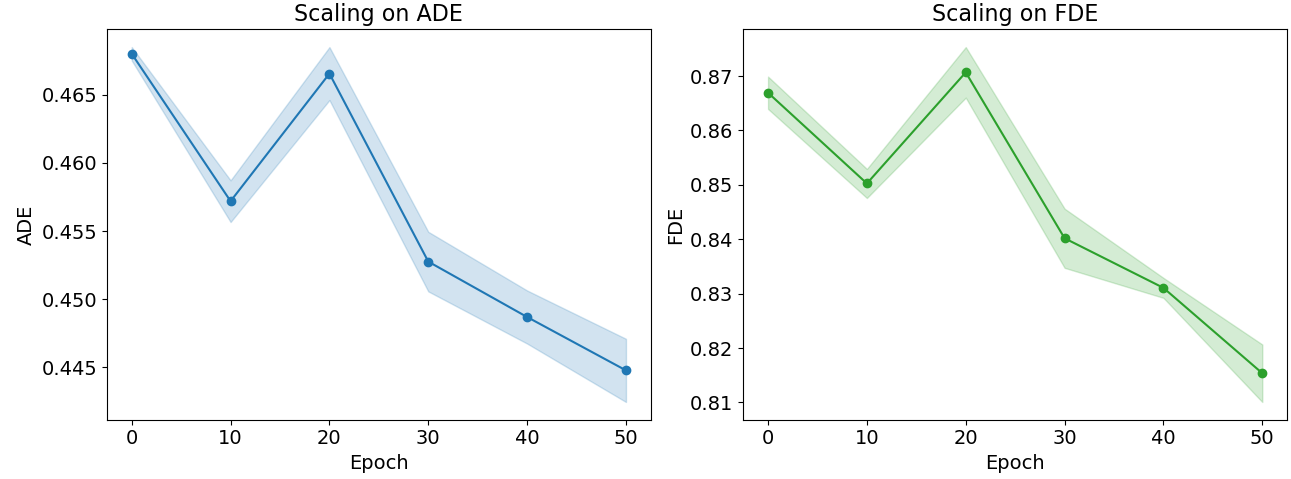}
    \caption{Scaling performance of RNTransformer.}\label{fig:scaling}
\end{figure}

We also examined the scaling properties of the Transformer \citep{zhai2022scaling} using RNTransformer pre-trained over different epochs. Figure \ref{fig:scaling} illustrates the transitions of ADE and FDE across different epochs. The results clearly show a consistent trend as the RNTransformer undergoes more training, the performance of the overall model steadily improves. This improvement is evident in the gradual reduction of both ADE and FDE metrics and proved the enhanced accuracy and reliability of pedestrian trajectory predictions with extended training.

These findings raise several pertinent questions for future research. Firstly, it would be valuable to explore the performance improvement associated with extended training. Identifying the saturation point of the model remains the essential question. Moreover, due to computation limitations, we were unable to find the optimal number of grids for forming the road network. The granularity of the grids impacts the model's capability to understand the spatial representation to enhance the model's performance. In addition to these considerations, exploring the effectiveness of trip modality across various types of trajectory prediction models is important. While our current work has demonstrated the utility of trip modality in socially aware pedestrian trajectory prediction, its applicability to non-socially aware models and vehicle trajectory prediction models warrants further investigation.

\section{Conclusions}

This study introduces a novel approach to enhance pedestrian trajectory prediction by incorporating trip information as a new modality. We propose RNTransformer to capture the global contextual trip information by reasoning pedestrian intention with crowd behavior. The RNTransformer significantly enhances prediction accuracy when integrating with various socially aware local pedestrian prediction models across multiple datasets. By formulating the scene into multiple grids and aggregating the number of pedestrians at each grid through time steps, the RNTransformer effectively learns crowd behavior. Moreover, the integrated model considers the impact of road environments on predictions by learning the spatial dependencies with GCN and temporal dependencies with GRU, as well as the pedestrian trajectories with the local model.

The proposed model is evaluated using pedestrian trajectories obtained from the ETH, UCY, and SDD datasets. Results demonstrate the quantitative improvements of our model that validate the effectiveness of incorporating trip information. Furthermore, the RNTransformer provides effective guidance for local models, resulting in more accurate goal predictions and better alignment with actual pedestrian trajectories. Future research should focus on expanding our methodology by exploring various strategies for sampling the grids in the road network and determining optimal scaling parameters. These efforts will help synthesize more accurate crowd behaviors and further enhance the predictive capabilities of the model. Additionally, we envision incorporating alternative approaches, such as non-socially aware models and vehicle trajectory predictions, to account for trip modality's effectiveness.


\bibliographystyle{elsarticle-harv}


\bibliography{reference}

\end{document}